\documentclass[authoryear]{ifacconf}

\usepackage{graphicx} % include this line if your document contains figures
\usepackage{balance}

\makeatletter
\let\old@ssect\@ssect % Store how ifacconf defines \@ssect
\makeatother

\usepackage{natbib}
\usepackage{hyperref}

\makeatletter
\def\@ssect#1#2#3#4#5#6{%
  \NR@gettitle{#6}% Insert key \nameref title grab
  \old@ssect{#1}{#2}{#3}{#4}{#5}{#6}% Restore ifacconf's \@ssect
}
\makeatother

\usepackage{tikz}
\usetikzlibrary{fit,patterns,shapes,arrows.meta,matrix,
                chains,positioning,shapes.geometric,
                arrows,svg.path,decorations.pathreplacing,calc}
\pgfmathsetmacro{\cubex}{0.5}
\pgfmathsetmacro{\cubey}{2.5}
\pgfmathsetmacro{\cubez}{2.5}
\usepackage{multirow}
\usepackage{subfigure}
\usepackage{comment}
\usepackage{soul}
\begin{document}
\begin{frontmatter}

\title{Semantic Image Segmentation with Deep Learning for Vine Leaf Phenotyping} %\thanksref{footnoteinfo}} 

%\thanks[footnoteinfo]{Sponsor and financial support acknowledgment
%goes here. Paper titles should be written in uppercase and lowercase
%letters, not all uppercase.}

\author[First]{Petros N. Tamvakis} 
\author[First]{Chairi Kiourt}
\author[Second]{Alexandra D. Solomou}
\author[First]{George Ioannakis}
\author[First]{Nestoras C. Tsirliganis}

\address[First]{Athena Research and Innovation Center, 
   Xanthi, 67100 Greece (e-mail: [petros.tamvakis,chairiq,gioannak,tnestor]@athenarc.gr)}
\address[Second]{Institute of Mediterranean \& Forest Ecosystems, Hellenic Agricultural Organization "DEMETER",
Athens, 11528 Greece (e-mail: solomou@fria.gr)}
%\address[Third]{Athena Research and Innovation Center, 
%  Xanthi, 67100 Greece (e-mail: chairiq@athenarc.gr)}

\begin{abstract} % Abstract of not more than 250 words.

Plant phenotyping refers to a quantitative description of the plant’s properties, however in image-based phenotyping analysis, our focus is primarily on the plant's anatomical, ontogenetical and physiological properties. This technique reinforced by the success of Deep Learning in the field of image based analysis is applicable to a wide range of research areas making high-throughput screens of plants possible, reducing the time and effort needed for phenotypic characterization. In this study, we use Deep Learning methods (supervised and unsupervised learning based approaches) to semantically segment grapevine leaves images in order to develop an automated object detection (through segmentation) system for leaf phenotyping which will yield information regarding their structure and function. In these directions we studied several deep learning approaches with promising results as well as we reported some future challenging tasks in the area of precision agriculture. Our work contributes to plant lifecycle monitoring through which dynamic traits such as growth and development can be captured and quantified, targeted intervention and selective application of agrochemicals and grapevine variety identification which are key prerequisites in sustainable agriculture.

%Furthermore, the ability to automatically monitor agricultural fields with high precision is an important capability and a key prerequisite for targeted intervention and selective application of agro-chemicals that may prove to be a game-changer in sustainable agriculture.

%Recent increases in computational power in addition to the expansion of system memory capabilities have led to the development of Machine Learning-related Artificial Intelligence. Nowadays, computers can perform millions of calculations in matter of seconds using methodologies and models that consist of numerous processing layers stacked on top of each other in architectures that we call Deep Neural Networks. Deep Learning has attracted attention not only in numerous research fields but also in nearly almost every aspect of everyday life while applications that make use of it increase at an exponential rate with the majority of them aiming to deliver quick, automated and accurate services in image recognition \citep{cnn}, natural language processing, self-driving vehicles etc.
\end{abstract}

\begin{keyword}
Semantic segmentation, Grapevines, Phenotyping, Pattern recognition
\end{keyword}

\end{frontmatter}
%===============================================================================

\section{INTRODUCTION}
\begin{figure}
    \centering
    \includegraphics[width=\columnwidth,trim=2cm 2cm 0cm 0cm,] {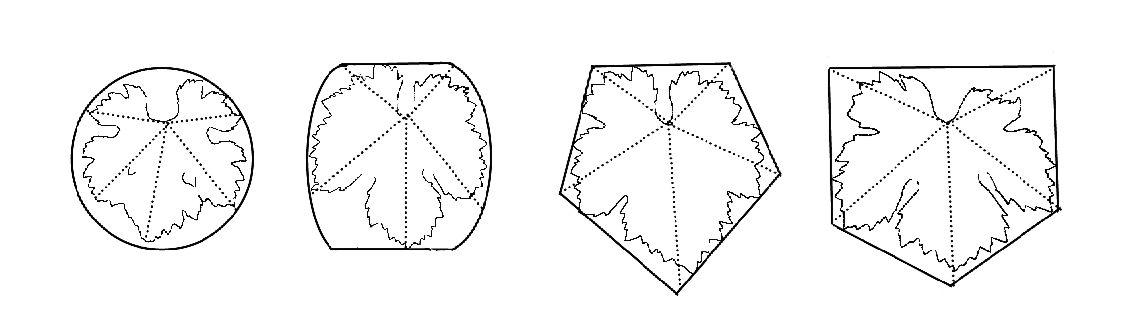}
    \caption{Different leaf blade shapes.}
    \label{fig:blade_shapes}
\end{figure}

\begin{figure}[ht]
  \centering
  \subfigure[Leaf dimensions]{\label{fig:leaf_dims}\includegraphics[width=0.3\columnwidth,height=50pt]{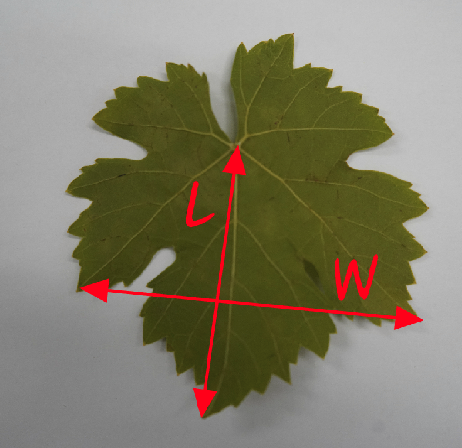}}
  \subfigure[Leaf parts]{\label{fig:blade}\includegraphics[width=0.4\columnwidth, height=50pt]{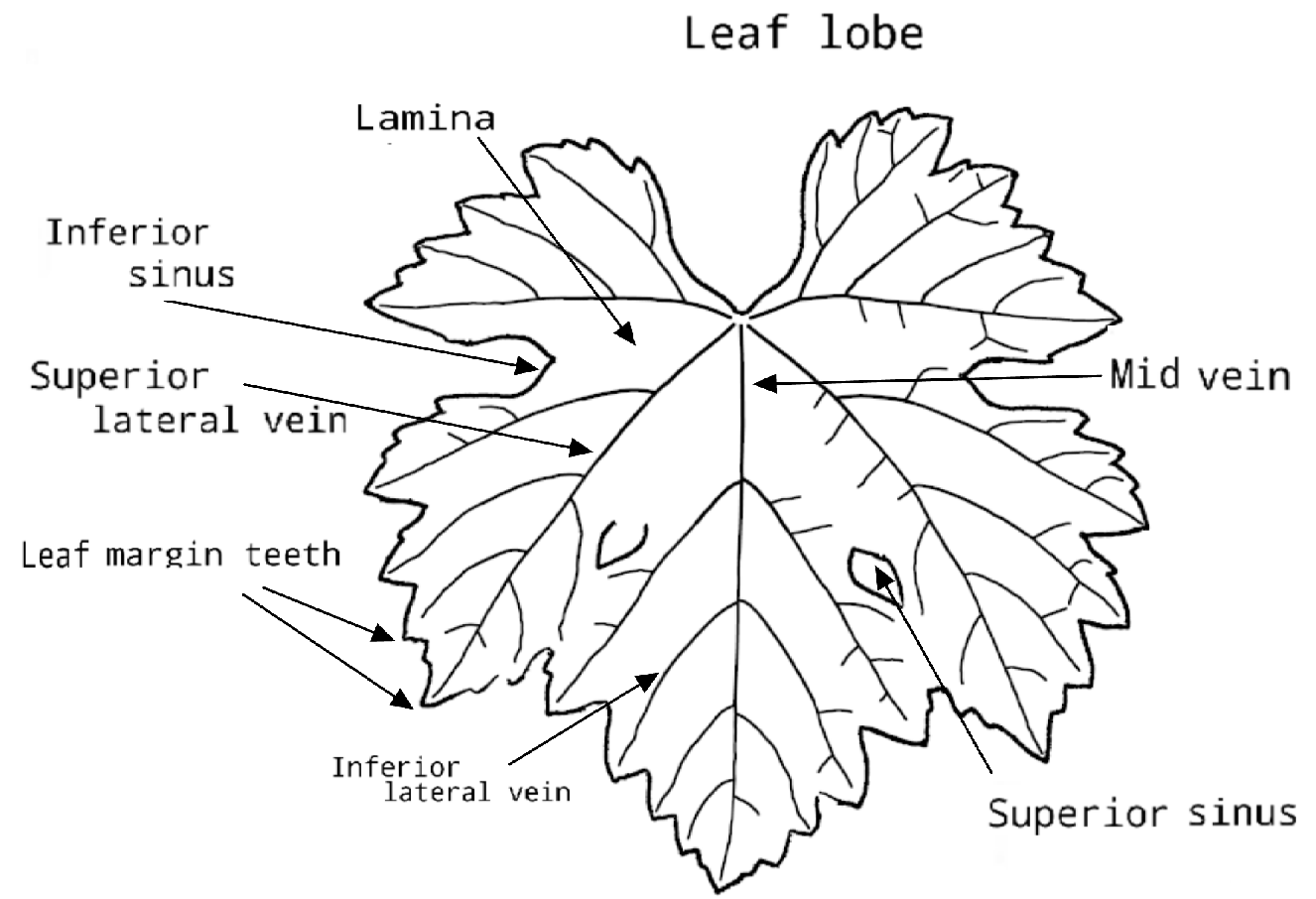}}
    \caption{Vine leaf characteristics.}
  \label{fig:leaf_attrs}
\end{figure}

Recently, in the wine industry, there is growing interest in exploring the role and effect that grapevine varieties play in adaptation strategies against global warming \citep{gutierrez}. In light of this, there is a need to ensure the genuiness of the plants in order to avoid planting the wrong material, which can result in considerable losses to viticulturists. The field of ampelography is concerned with identifying and classifying grapevine varieties using several parameters or descriptors \citep{soldavini,garcia}. It provides relevant morphological and agronomical information for varietal characterization studies, breeding programs and conservation purposes \citep{khalil}. It is possible to distinguish grapevine varieties by the phenotypic features of their leaves, which are unique to each variety \citep{galet}. More specifically, the features of the leaves are the following:  geometrical shape of the leaf surface,  perimeter of the leaf surface \citep{gamboa}, number of lobes, size of teeth, length of teeth, ratio length/width of teeth, shape of blade (Fig. \ref{fig:blade_shapes}), leaf length (L) and width (W) measurement (Fig. \ref{fig:leaf_dims}) \citep{eftekhari}, color of the upper side of blade, undulation of blade between main and lateral veins, general shape of petiole sinus, tooth at petiole sinus, petiole sinus limited by veins, shape of upper lateral sinus, depth of upper lateral sinus, shape of base, length of petiole compared to middle vein \citep{ipgri}. Grapevine mature leaf’s parts are illustrated in Fig. \ref{fig:blade}.

\par
Quantifying the dimensions of leaf veins and their structure has primarily relied on optical observation and empirical experience of domain experts i.e. agronomists, agriculturists, botanologists etc. However, due to the significantly high degree of complexity of vein networks, their intense color similarity with the overall leaf and the large number of various cultivated plants and their problems, even well trained experts of the area often fail to annotate successfully the over structure of the lead and are consequently led to mistaken conclusions.

%\begin{figure}[t]
%    \centering
%    \includegraphics[width=0.8\columnwidth, height=2.5cm, keepaspectratio]{photos/leaf_dims_cropped.png}
%    \caption{Grapevine leaves showing the point of leaf length (L) and width (W) measurement.}
%    \label{fig:leaf_dims}
%\end{figure}

Unfortunately, there exists no suitable empirical method to quantify physical vein network geometry with sufficient scope and resolution which makes the evaluation of the predicted vein network structure nearly impossible \citep{price}. The development of an automated computational system that detects leaf vein network structure, in real-time, would offer a valuable tool in the hands of the agronomist who wishes to analyze such networks and formulate/evaluate hypotheses regarding their structure and function. Furthermore it can aid agricultural robots: unmanned ground and aerial vehicles will identify genetic traits, crop growth and plant diseases. Agribots equipped with such systems will observe and measure crop growth and inform on the fly about a plant's performance against its predicted growth plan. In addition, the process of the automated leaf samples collection, plant harvesting and pruning through soft robotic arms (fingers), will be strengthened with more accurate and targeted actions.
\par

This need spurred the development of a number of theoretical Machine Learning (ML) models that predict optimal vein network structures across a broad array of taxonomic groups, from mammals to plants. In particular, a number of empirical methods to quantify the vein network structure of plants, from roots, to xylem networks in shoots and within leaves have been developed. 

%\begin{figure}[t]
%    \centering
%    \includegraphics[width=0.8\columnwidth, height=3cm, keepaspectratio]{photos/leaf_parts3_cropped.png}
%    \caption{Mature leaf parts.}
%    \label{fig:blade}
%\end{figure}

\par
Advances in automated plant handling and digital imaging along with the recent increases in computational power and the expansion of system memory capabilities now make it possible to use image-based analysis for plant phenotyping that allows for high-throughput characterization of the extracted plant traits and improved understanding of the adaptive and ecological significance of vessel bundle (veins) network structure.

\begin{comment}
The rest of the paper is organised as follows. The following section provides a brief literature review of the domain. The third section covers the methods, algorithms and dataset used in the experiments. The fourth section discusses the outcomes of the experiments, while the last section concludes the paper by summarising the key points and sets the future directions.
\end{comment}

\section{RELATED WORK}

The introduction of Deep Learning (DL) techniques into agriculture \citep{carranza} began to take place in the context of precision agriculture. Traditional ML approaches that have been extensively adopted in the agricultural field, such as plant disease investigation and pest detection are gradually being replaced by more sophisticated techniques.The majority of which rely on some kind of image recognition/classification i.e. DL image processing, image-based phenotyping, including weed detection \citep{milioto}, crop disease diagnosis \citep{mohanty,bresilla}, fruit detection \citep{ghosal}, and many other applications as listed in the recent review \citep{kamilaris}. 
\par
The notion that leaves are a major hydraulic bottleneck in plants \citep{sack}, motivated attempts to model patterns of conductance \citep{cochard,price3} by measuring the geometry of veins. Vein network structure can influence photosynthesis via hydraulic efficiency, with recent work implicating vein density as a good predictor of photosynthetic rates \citep{brodribb,sack2}. Leaf vein patterning is also associated with leaf shape in general, suggesting shared developmental pathways \citep{dengler2}. In \cite{price2}, the authors developed a tool which produces descriptive statistics about the dimensions and positions of leaf veins and areoles by utilizing a series of thresholding, cleaning, and segmentation algorithms applied to images of leaf veins. An all‐inclusive software tool for mathematical and statistical calculations in plant growth analysis that calculates up to six of the most fundamental growth parameters according to a purely classical approach across one harvest‐interval was developed in \citep{Hunt}. Today the field has broadened its range from the initial characterization of single-plant traits in controlled conditions towards real-life applications of robust field techniques in plant plots and canopies \citep{walter}. 
\par
Convolutional Neural Networks (CNN) \citep{cnn} have been used in image recognition with remarkable success and constitute one of the most powerful DL techniques for modeling complex processes such as pattern recognition in image based applications. Naturally, the majority of image-based approaches in precision agriculture are based on popular CNN architectures. In \citep{lee} the authors developed CNN architectures for automatic leaf-based plant recognition. \cite{pawara} compared the performance of some conventional pattern recognition techniques with that of CNN models, in plant identification, using three different image-databases of either entire plants and fruits, or plant leaves, concluding that CNNs drastically outperform conventional methods. \cite{grinblat} presented a simple, yet powerful Neural Network (NN) for the successful identification of three different legume species based on the morphological patterns of leave nerves. In their work, \cite{dyrmann} presented a method that can recognize weeds and plant species using colored images. They used CNNs and tested a total 10.413 images of 22 weeds and crop species. Their model was able to achieve a classification accuracy of 86.2\%. \cite{mohanty} developed a smartphone-assisted disease diagnosis system by employing two state-of-the-art architectures: AlexNet \citep{alexnet} and GoogLeNet \citep{szegedy} and trained their model to identify 14 crop species and 26 diseases.

\section{METHOD}

Semantic image segmentation \citep{ronneberger,he,long} is a popular image analyzing technique where each pixel is assigned to one from a set of predefined classes. It is a complex method that entails the description, categorization, and visualization of the regions of interest in an image contributing to complete scene understanding. Semantic segmentation models first determine the presence or not of the objects of interest in the picture and then classify object's pixels to their corresponding class.

\subsection{Dataset}
At this part it should be highlighted that the data acquisition and dataset organization approach is targeted on small number of images. Our dataset consists of 24 images with dimensions 7,952x5,304 pixels of vine leaves that belong to cabernet chauvignon, mavroudi and merlot varieties, collected from an organic farm located in Alexandroupoli in Thrace, a region in northeastern Greece. Pictures of the collected leaves were taken by a Sony $\alpha7$ III camera model in a controlled environment, in laboratory under artificial lighting conditions. Collecting multiple unique and good quality images in organic vineyards is difficult, because they present multiple leaf damages. Natural lighting, variation in illumination, camera viewpoint, leaf clustering and wind conditions impede photo shooting even more. In addition to these reasons, for the Supervised Learning (SL) part of this study, dataset size was kept significantly small by design mainly because manual annotation of images is tedious and time consuming. Another, key objective was to examine the efficacy of a U-Net \citep{ronneberger} CNN architecture with as low available data as possible. For this reason, a small number of synthetic photos were acquired from the original ones by the use of image augmentation techniques (random rotation and random zoom) and repeated runs of increasing training/validation/test dataset size were examined (Table \ref{tab:dataset-results}). It is a well known fact that most DL-based image segmentation studies suffer from the lack of annotated data. Training a NN on a small dataset has downsides such as reduced generalizing capabilities, but on the other hand, the low level of scene complexity and the controlled conditions under which the pictures were collected, produces a high degree of feature similarity which compensates for the lack of dataset size.
\par

\begin{figure}[ht]
  \centering
  \includegraphics[width=0.24\columnwidth, height=42pt]{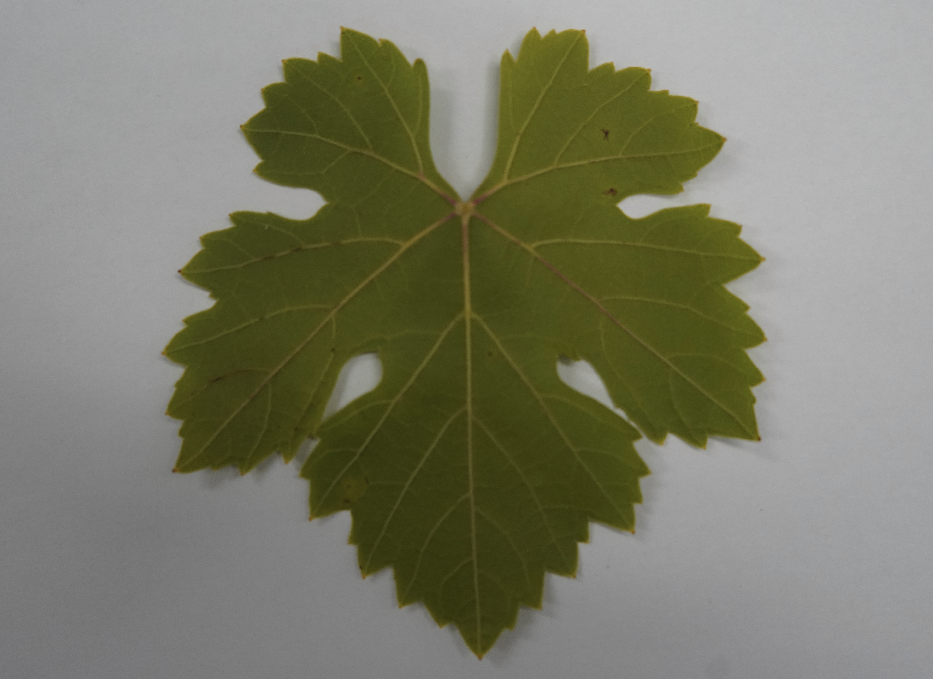}
  \hfill
  \includegraphics[width=0.24\columnwidth, height=42pt]{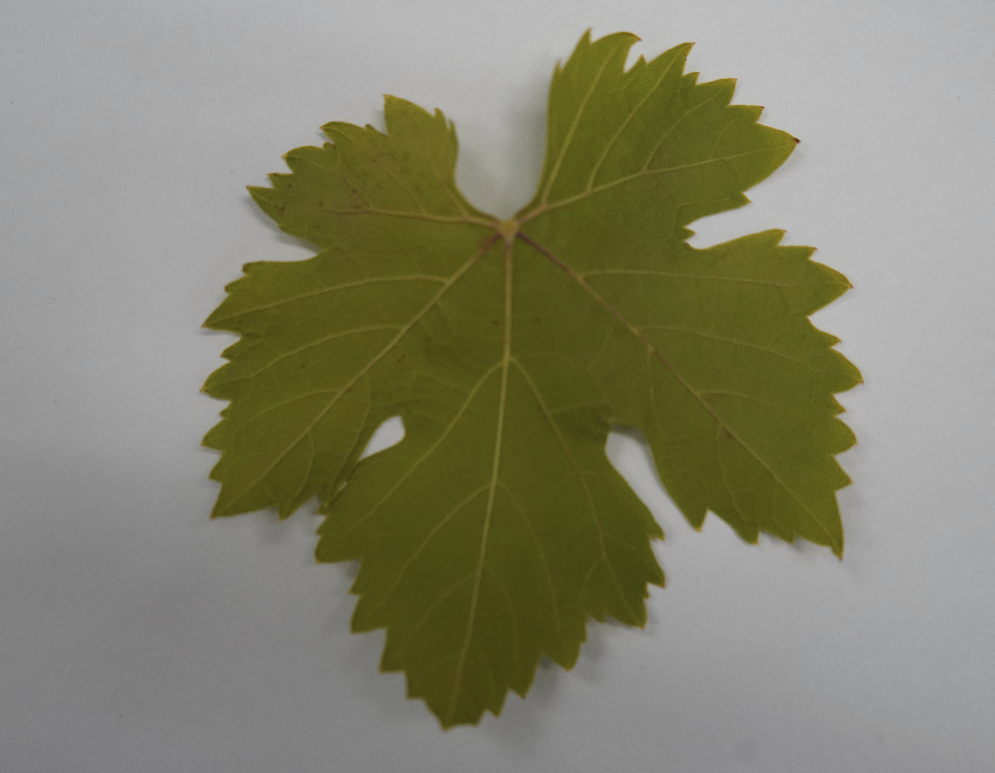}
  \hfill
  \includegraphics[width=0.24\columnwidth, height=42pt]{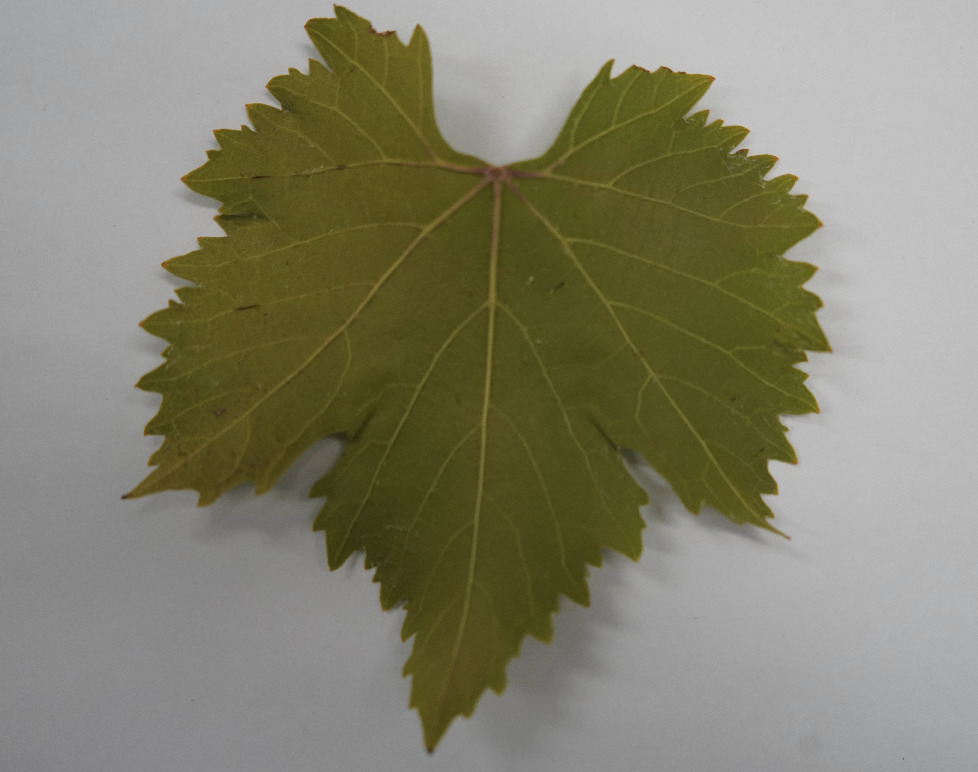}
  \hfill
  \includegraphics[width=0.24\columnwidth, height=42pt]{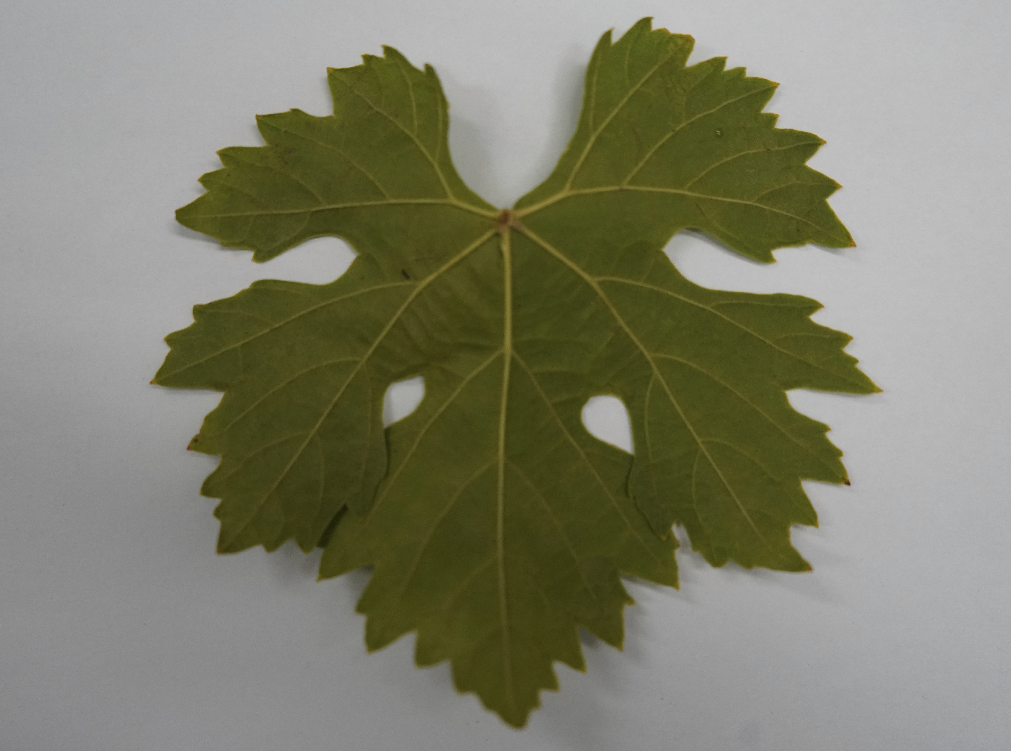}
  \vfill
  \includegraphics[width=0.24\columnwidth, height=42pt]{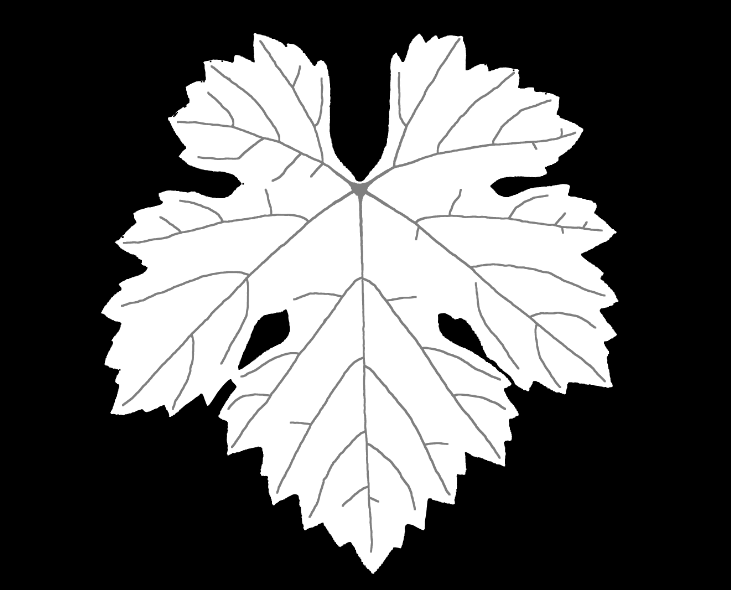}
  \hfill
  \includegraphics[width=0.24\columnwidth, height=42pt]{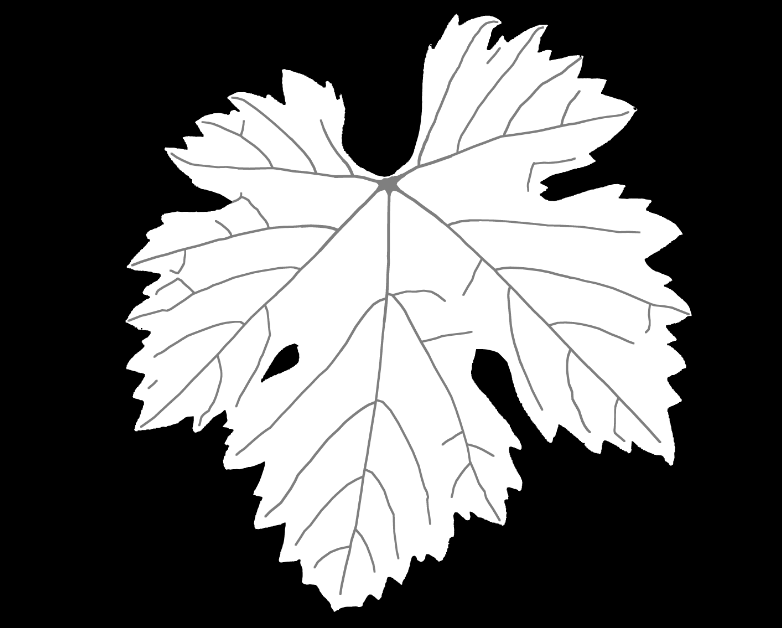}
  \hfill
  \includegraphics[width=0.24\columnwidth, height=42pt]{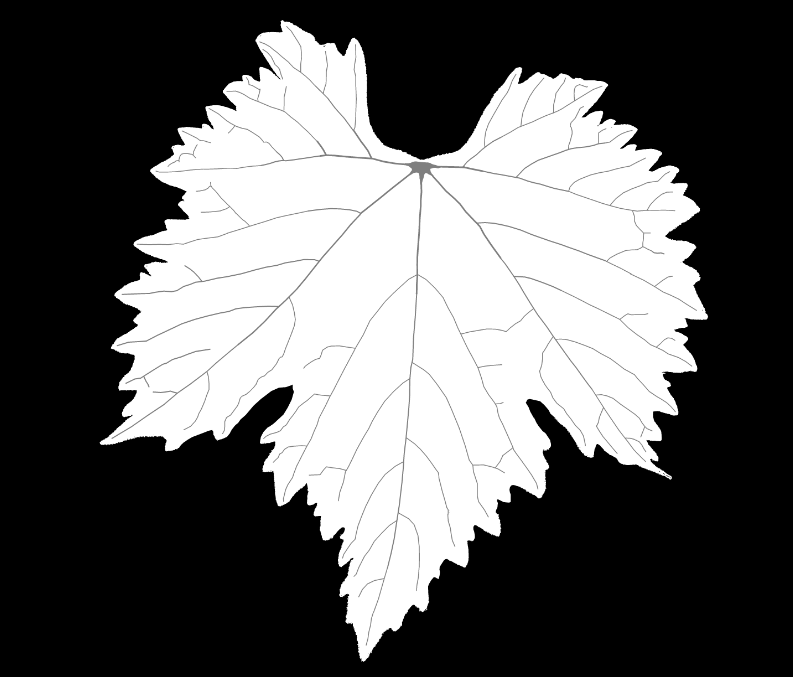}
  \hfill
  \includegraphics[width=0.24\columnwidth, height=42pt]{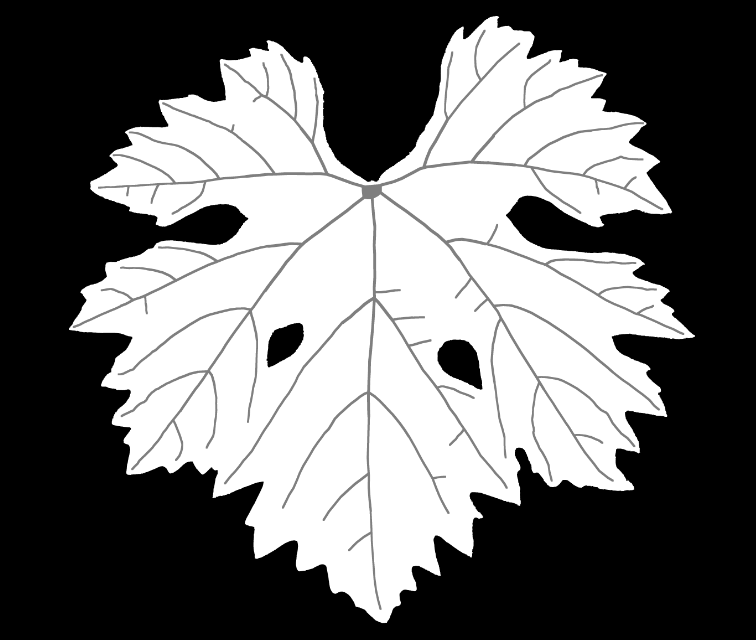}
  %\hfill
  
  \caption{Dataset samples (\textit{top figures}) and their corresponding trimaps (\textit{bottom figures}).}
  \label{fig:dataset_samples}
\end{figure}

This research focused on two phenotypic characteristics: leaf veins and leaf blade, each one representing a different class. Our models' task was to segment the images, singling out the aforementioned characteristics from the background. The dataset was annotated manually by the use of machine vision libraries such as OpenCV v4.5\citep{opencv} and GIMP v2.10 \citep{gimp} which is an open source image manipulation software, following field expert's guidelines and instructions, creating in this way for each image its corresponding trimap (Fig.\ref{fig:dataset_samples}). An automated annotation pipeline through computer vision approaches (noise removal, morphological transformations, binary thresholding and edge detection) was also developed but the resulting images lacked the level of precision of the manually annotated ones. 

\subsection{Supervised learning based semantic segmentation}
In this section we describe the adopted U-Net like, three different approaches, based on SL, which are:
\begin{itemize}
    \item design and train from scratch,
    \item transfer learning and
    \item parameterization of a popular architecture.
\end{itemize}

In our approaches the NN architectures are inspired and based on the typical U-Net architecture, each one with minor variations.

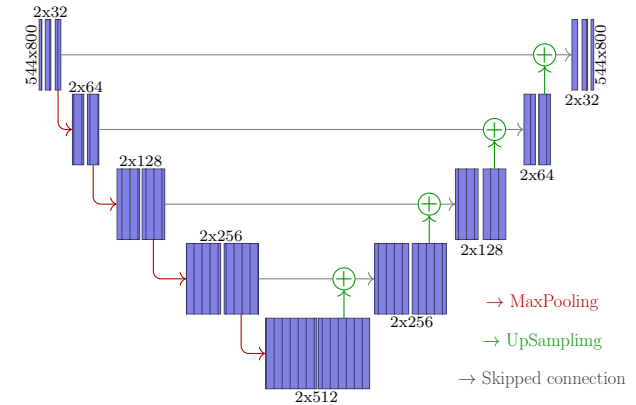
\begin{figure}[ht]
    \centering
    \begin{tikzpicture}[scale=0.33]

                    \tikzstyle{operator} = [
        			    circle,
        			    scale=0.6,
        			    draw,
        			    color=green!60!black!,
        			    inner sep=-0.5pt,
        			    minimum height =0.5cm,
        			    ];
        			    
        			\tikzstyle{concat} = [
        			    gray,
        			    ];
        			    
        			\tikzstyle{upsample} = [
        			    color=green!60!black,
        			    ];
        			    
    				\tikzstyle{filters} = [
    				    scale=0.6,
    				    rectangle,
        			    minimum width=2mm,
        			    minimum height=2mm,
        			    inner sep=1pt
        			    ];

    				\tikzstyle{layer} = [
    				    draw=black,
    				    scale=0.5,
    				    preaction={fill=blue!80!black},
    				    pattern=vertical lines,
    				    opacity=0.5,
    				    rectangle,
        			    minimum width=2mm,
        			    minimum height=25mm,
        			    inner sep=1pt
        			    ];
        		   
        		   \tikzstyle{my_arrow} = [
        		        color=red!70!black,
        		    	rounded corners=.1cm,
        			    ];
        			    
        		   \tikzstyle{filters} = [
    				    rectangle,
        			    minimum width=2mm,
        			    minimum height=2mm,
        			    inner sep=1pt
        			    ];
        			    
        	% layers		    
            \node [layer] [scale=0.75] (enc1_1) [minimum width=1mm] at (0,0) {};
            \node [layer] [scale=0.75] (enc1_2)  at 
                    ([xshift=({3 mm})]enc1_1){};
            \node [layer] [scale=0.75] (enc1_3)  at 
                    ([xshift=({4 mm})]enc1_2){};
            \node [layer] [scale=0.75] (enc2_1) [minimum width=4mm]  at
                    (1.5,-3) {};
            \node [layer] [scale=0.75](enc2_2) [minimum width=4mm]  at 
                    ([xshift=({6 mm})]enc2_1){};
            \node [layer] [scale=0.75](enc3_1) [minimum width=8mm]  at
                    (3.5,-6) {};
            \node [layer] [scale=0.75](enc3_2) [minimum width=8mm]  at 
                    ([xshift=({10 mm})]enc3_1){};
            \node [layer] [scale=0.75](enc4_1) [minimum width=12mm]  at
                    (6.5,-9) {};
            \node [layer] [scale=0.75](enc4_2) [minimum width=12mm]  at 
                    ([xshift=({15 mm})]enc4_1){};
            \node [layer] [scale=0.75](enc5_1) [minimum width=18mm]  at
                    (10,-12) {};
            \node [layer] [scale=0.75](enc5_2) [minimum width=18mm]  at 
                    ([xshift=({21 mm})]enc5_1){};
            \node [layer] [scale=0.75](dec1_1) [minimum width=12mm]  at
                    (14,-9) {};
            \node [layer] [scale=0.75](dec1_2) [minimum width=12mm]  at 
                    ([xshift=({15 mm})]dec1_1){};
            \node [layer] [scale=0.75](dec2_1) [minimum width=8mm]  at
                    (17,-6) {};
            \node [layer] [scale=0.75](dec2_2) [minimum width=8mm]  at 
                    ([xshift=({11 mm})]dec2_1){};
            \node [layer] [scale=0.75](dec3_1) [minimum width=4mm]  at
                    (19.5,-3) {};
            \node [layer] [scale=0.75](dec3_2) [minimum width=4mm]  at 
                    ([xshift=({6 mm})]dec3_1){};
            \node [layer] [scale=0.75](dec4_1) at (21.3,0) {};
            \node [layer] [scale=0.75](dec4_2) at 
                    ([xshift=({4 mm})]dec4_1){};
            \node [layer] [scale=0.75](dec4_3) [minimum width=1mm] at 
                    ([xshift=({3 mm})]dec4_2){};
            
            % addition operators
            \node [operator] [scale=0.75](add1) at (20.1,0) {\textbf{\Large{+}}};
            \node [operator] [scale=0.75](add2) at (18.1,-3) {\textbf{\Large{+}}};
            \node [operator] [scale=0.75](add3) at (15.5,-6) {\textbf{\Large{+}}};
            \node [operator] [scale=0.75](add4) at (12.1,-9) {\textbf{\Large{+}}};

            % arrows and skipped connections
            \draw [upsample] [->] (dec3_2.north) -- (add1.south);
            \draw [upsample] [->] (dec2_2.north) -- (add2.south);
            \draw [upsample] [->] (dec1_2.north) -- (add3.south);
            \draw [upsample] [->] (enc5_2.north) -- (add4.south);
            
            \draw [concat] (enc1_3.east) -- (add1.west);
            \draw [concat] (enc2_2.east) -- (add2.west);
            \draw [concat] (enc3_2.east) -- (add3.west);
            \draw [concat] (enc4_2.east) -- (add4.west);
            
            \draw [concat] [->] (add1.east) -- (dec4_1.west) ;
            \draw [concat] [->] (add2.east) -- (dec3_1.west);
            \draw [concat] [->] (add3.east) -- (dec2_1.west);
            \draw [concat] [->] (add4.east) -- (dec1_1.west);
            
            \draw [my_arrow] [->](enc1_3.south) |- (enc2_1.west);
            \draw [my_arrow] [->](enc2_2.south) |- (enc3_1.west);
            \draw [my_arrow] [->](enc3_2.south) |- (enc4_1.west);
            \draw [my_arrow] [->](enc4_2.south) |- (enc5_1.west);
            
            % filters description
            \node [filters] [scale=0.75] at ([yshift=({3mm}), xshift=({1mm})]enc1_2.north)
                {\footnotesize{2x32}};
            \node [filters] [scale=0.75] at ([yshift=({3mm}), xshift=({3mm})]enc2_1.north)
                {\footnotesize{2x64}};
            \node [filters] [scale=0.75] at ([yshift=({3mm}), xshift=({5mm})]enc3_1.north)
                {\footnotesize{2x128}};
            \node [filters] [scale=0.75] at ([yshift=({3mm}), xshift=({7mm})]enc4_1.north)
                {\footnotesize{2x256}};
            \node [filters] [scale=0.75] at ([yshift=({-3mm}), xshift=({10mm})]enc5_1.south)
                {\footnotesize{2x512}};
            \node [filters] [scale=0.75] at ([yshift=({-4mm}), xshift=({3mm})]dec4_1.south)
                {\footnotesize{2x32}};
            \node [filters] [scale=0.75] at ([yshift=({-4mm}), xshift=({3mm})]dec3_1.south)
                {\footnotesize{2x64}};
            \node [filters][scale=0.75] at ([yshift=({-4mm}), xshift=({6mm})]dec2_1.south)
                {\footnotesize{2x128}};
            \node [filters] [scale=0.75] at ([yshift=({-3mm}), xshift=({8mm})]dec1_1.south)
                {\footnotesize{2x256}};
            
            % input-output description
            \node [filters] [scale=0.75, rotate=90] at ([xshift=({-3mm})]enc1_1.west)
                {\footnotesize{544x800}};
            \node [filters] [scale=0.75, rotate=90] at ([xshift=({4mm})]dec4_3.west)
                {\footnotesize{544x800}};
                
            % the legend
          \matrix [row sep=2mm] at (20,-11.5) {
              \node [scale=0.40] {\Large{\color{red!70!black} $\rightarrow$ MaxPooling}};\\
              \node [upsample] [scale=0.40] {\Large{\color{green!60!black} $\rightarrow$ UpSamplimg}};\\
              \node [concat] [scale=0.40] {\Large{\color{gray!50!black} $\rightarrow$ Skipped connection}};\\
        };
                
    \end{tikzpicture}
    \caption{U-Net architecture (input size, number of blocks and filters differ depending on the application).}
    \label{fig:unet_arch}
\end{figure}

For the \(1^{st}\) approach images are downsized to 544x800 pixels. The network implements the skipped connections between the downsampling and upsampling layer stacks and consists of double 32, 64, 128, 256, and 512-filtered Convolutional layer blocks. We used the Adam optimizer \citep{adam} with a fixed learning rate of $10^{-3}$. The decoder part on the other hand is symmetrical to its contracting counterpart (Fig.\ref{fig:unet_arch}). 

The same NN architecture through the transfer learning approach was tested in an attempt to examine if the use of pre-trained weights will yield better results (\(2^{nd}\) approach). We exploited the MobileNetV2 \citep{mobilenetv2} as the encoder backbone which was pre-trained on ImageNet \citep{imagenet}. The two models (\(1^{st}\) and \(2^{nd}\) approaches) differ in the input dimensions: for the second approach the original input (images sizes) are downsized to 224x224 pixels so as to fit the pre-trained weights.
\par
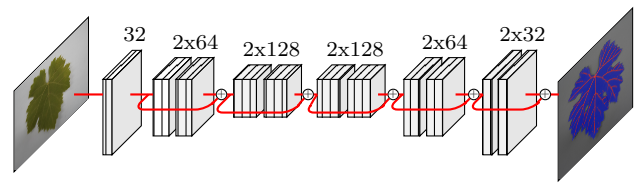
\begin{figure}[ht]
    \centering
    \begin{tikzpicture}[scale=0.40]
    
                    \tikzstyle{operator} = [
        			    scale=0.75, 
        			    circle,
        			    draw,
        			    inner sep=-0.5pt,
        			    minimum height =.2cm,
        			    ];
        			\tikzstyle{flow} = [
        			    thick, 
        			    red, 
        			    rounded corners
        			    ];
        			\tikzstyle{photo} = [
        			    scale=0.8,
        			    transform shape,
        			    draw,
        			    black,
        			    inner sep=0.2mm
        			    ];
    				\tikzstyle{filters} = [
    				    rectangle,
        			    minimum width=2mm,
        			    minimum height=2mm,
        			    inner sep=1pt
        			    ];

        \draw[draw=black, preaction={fill=black!5}, pattern=vertical lines] (0,0,0) -- ++(-\cubex+0.2,0,0) -- ++(0,-\cubey,0) -- ++(\cubex-0.2,0,0) -- cycle;
        \draw[draw=black,fill=black!10] (0,0,0) -- ++(0,0,-\cubez) -- ++(0,-\cubey,0) -- ++(0,0,\cubez) -- cycle;
        \draw[draw=black,fill=black!5, pattern=north east lines] (0,0,0) -- ++(-\cubex+0.2,0,0) -- ++(0,0,-\cubez) -- ++(\cubex-0.2,0,0) -- cycle;

        \draw[draw=black,fill=black!5, scale=0.75, pattern=vertical lines] (2.5,0,0) -- ++(-\cubex-0.2,0,0) -- ++(0,-\cubey,0) -- ++(\cubex+0.2,0,0) -- cycle;
        \draw[draw=black,fill=black!10, scale=0.75] (2.5,0,0) -- ++(0,0,-\cubez) -- ++(0,-\cubey,0) -- ++(0,0,\cubez) -- cycle;
        \draw[draw=black,fill=black!5, scale=0.75, pattern=north east lines] (2.5,0,0) -- ++(-\cubex-0.2,0,0) -- ++(0,0,-\cubez) -- ++(\cubex+0.2,0,0) -- cycle;
        
        \draw[draw=black, preaction={fill=black!5}, scale=0.75, pattern=vertical lines] (3.5,0,0) -- ++(-\cubex-0.2,0,0) -- ++(0,-\cubey,0) -- ++(\cubex+0.2,0,0) -- cycle;
        \draw[draw=black,fill=black!10, scale=0.75] (3.5,0,0) -- ++(0,0,-\cubez) -- ++(0,-\cubey,0) -- ++(0,0,\cubez) -- cycle;
        \draw[draw=black, preaction={fill=black!5}, scale=0.75, pattern=north east lines] (3.5,0,0) -- ++(-\cubex-0.2,0,0) -- ++(0,0,-\cubez) -- ++(\cubex+0.2,0,0) -- cycle;
        
        \draw[draw=black, preaction={fill=black!5}, pattern=vertical lines, scale=0.5] (9.5,0,0) -- ++(-\cubex-1,0,0) -- ++(0,-\cubey,0) -- ++(\cubex+1,0,0) -- cycle;
        \draw[draw=black, fill=black!10, scale=0.5] (9.5,0,0) -- ++(0,0,-\cubez) -- ++(0,-\cubey,0) -- ++(0,0,\cubez) -- cycle;
        \draw[draw=black,preaction={fill=black!5}, pattern=north east lines, scale=0.5] (9.5,0,0) -- ++(-\cubex-1,0,0) -- ++(0,0,-\cubez) -- ++(\cubex+1,0,0) -- cycle;

        \draw[draw=black, preaction={fill=black!5}, pattern=vertical lines, scale=0.5] (11.5,0,0) -- ++(-\cubex-1,0,0) -- ++(0,-\cubey,0) -- ++(\cubex+1,0,0) -- cycle;
        \draw[draw=black,fill=black!10, scale=0.5] (11.5,0,0) -- ++(0,0,-\cubez) -- ++(0,-\cubey,0) -- ++(0,0,\cubez) -- cycle;
        \draw[draw=black, preaction={fill=black!5}, pattern=north east lines, scale=0.5] (11.5,0,0) -- ++(-\cubex-1,0,0) -- ++(0,0,-\cubez) -- ++(\cubex+1,0,0) -- cycle;
        
        \draw[draw=black, preaction={fill=black!5}, pattern=vertical lines, scale=0.5] (15,0,0) -- ++(-\cubex-1,0,0) -- ++(0,-\cubey,0) -- ++(\cubex+1,0,0) -- cycle;
        \draw[draw=black,fill=black!10, scale=0.5] (15,0,0) -- ++(0,0,-\cubez) -- ++(0,-\cubey,0) -- ++(0,0,\cubez) -- cycle;
        \draw[draw=black,, preaction={fill=black!5}, pattern=north east lines,scale=0.5] (15,0,0) -- ++(-\cubex-1,0,0) -- ++(0,0,-\cubez) -- ++(\cubex+1,0,0) -- cycle;

        \draw[draw=black, preaction={fill=black!5}, pattern=vertical lines, scale=0.5] (17,0,0) -- ++(-\cubex-1,0,0) -- ++(0,-\cubey,0) -- ++(\cubex+1,0,0) -- cycle;
        \draw[draw=black,fill=black!10, scale=0.5] (17,0,0) -- ++(0,0,-\cubez) -- ++(0,-\cubey,0) -- ++(0,0,\cubez) -- cycle;
        \draw[draw=black, preaction={fill=black!5}, pattern=north east lines, scale=0.5] (17,0,0) -- ++(-\cubex-1,0,0) -- ++(0,0,-\cubez) -- ++(\cubex+1,0,0) -- cycle;
        
        \draw[draw=black,, preaction={fill=black!5}, pattern=vertical lines, scale=0.75] (13.5,0,0) -- ++(-\cubex-0.2,0,0) -- ++(0,-\cubey,0) -- ++(\cubex+0.2,0,0) -- cycle;
        \draw[draw=black,fill=black!10, scale=0.75] (13.5,0,0) -- ++(0,0,-\cubez) -- ++(0,-\cubey,0) -- ++(0,0,\cubez) -- cycle;
        \draw[draw=black, preaction={fill=black!5}, pattern=north east lines, scale=0.75] (13.5,0,0) -- ++(-\cubex-0.2,0,0) -- ++(0,0,-\cubez) -- ++(\cubex+0.2,0,0) -- cycle;
        
        \draw[draw=black, preaction={fill=black!5}, pattern=vertical lines, scale=0.75] (14.5,0,0) -- ++(-\cubex-0.2,0,0) -- ++(0,-\cubey,0) -- ++(\cubex+0.2,0,0) -- cycle;
        \draw[draw=black,fill=black!10, scale=0.75] (14.5,0,0) -- ++(0,0,-\cubez) -- ++(0,-\cubey,0) -- ++(0,0,\cubez) -- cycle;
        \draw[draw=black, preaction={fill=black!5}, pattern=north east lines, scale=0.75] (14.5,0,0) -- ++(-\cubex-0.2,0,0) -- ++(0,0,-\cubez) -- ++(\cubex+0.2,0,0) -- cycle;

        \draw[draw=black, preaction={fill=black!5}, pattern=vertical lines,] (12.5,0,0) -- ++(-\cubex+0.2,0,0) -- ++(0,-\cubey,0) -- ++(\cubex-0.2,0,0) -- cycle;
        \draw[draw=black,fill=black!10] (12.5,0,0) -- ++(0,0,-\cubez) -- ++(0,-\cubey,0) -- ++(0,0,\cubez) -- cycle;
        \draw[draw=black, preaction={fill=black!5}, pattern=north east lines,] (12.5,0,0) -- ++(-\cubex+0.2,0,0) -- ++(0,0,-\cubez) -- ++(\cubex-0.2,0,0) -- cycle;

        \draw[draw=black, preaction={fill=black!5}, pattern=vertical lines,] (13,0,0) -- ++(-\cubex+0.2,0,0) -- ++(0,-\cubey,0) -- ++(\cubex-0.2,0,0) -- cycle;
        \draw[draw=black,fill=black!10] (13,0,0) -- ++(0,0,-\cubez) -- ++(0,-\cubey,0) -- ++(0,0,\cubez) -- cycle;
        \draw[draw=black, preaction={fill=black!5}, pattern=north east lines,] (13,0,0) -- ++(-\cubex+0.2,0,0) -- ++(0,0,-\cubez) -- ++(\cubex-0.2,0,0) -- cycle;
        
        %\node [operator][draw=white](add0) at (1.1,-0.5) {};
        \node [operator][scale=0.6](add1) at (3.6,-0.5) {+};
        \node [operator][scale=0.6](add2) at (6.45,-0.5) {+};
        \node [operator][scale=0.6](add3) at (9.25,-0.5) {+};
        \node [operator][scale=0.6](add4) at (11.9,-0.5) {+};
        \node [operator][scale=0.6](add5) at (14.25,-0.5) {+};
        
        \begin{scope}[cm={1.5, 1.5, 0, 2.25, (-2, -0.5)}]
            \node [photo] {
                    \includegraphics[width=2cm]{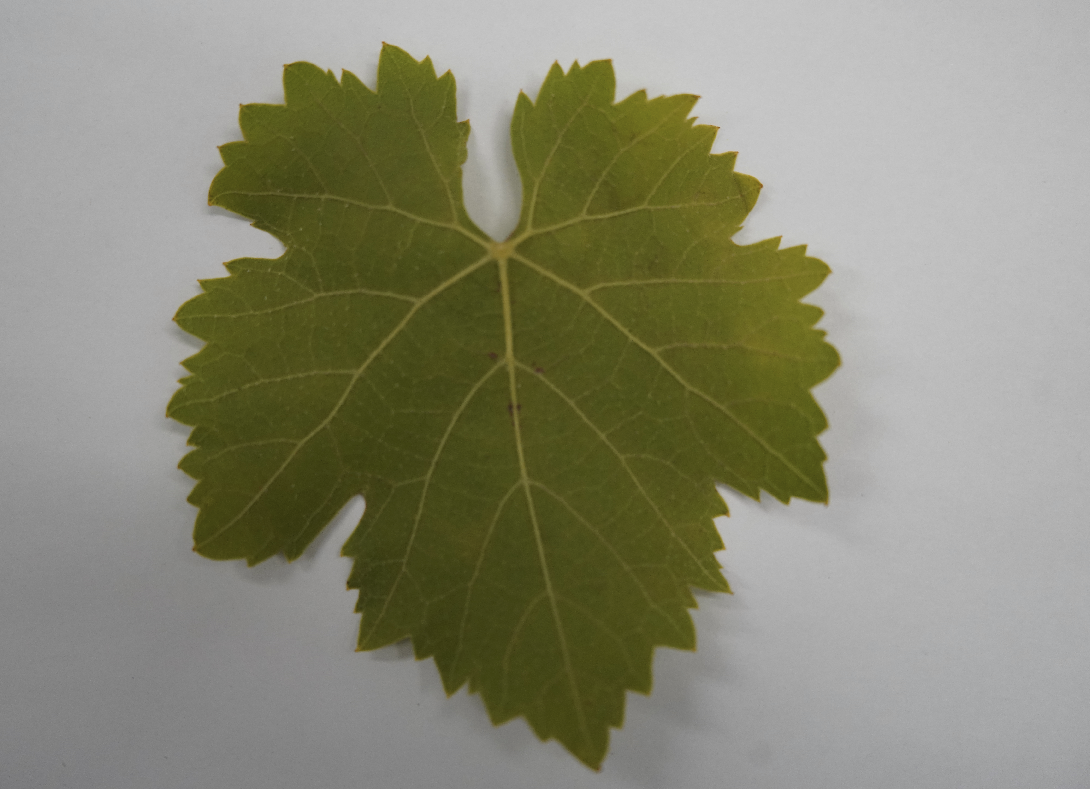}
                };
        \end{scope}

        \begin{scope}[cm={1.5, 1.5, 0, 2.25, (15.9, -0.5)}]
            \node[photo] {
                    \includegraphics[width=2cm]{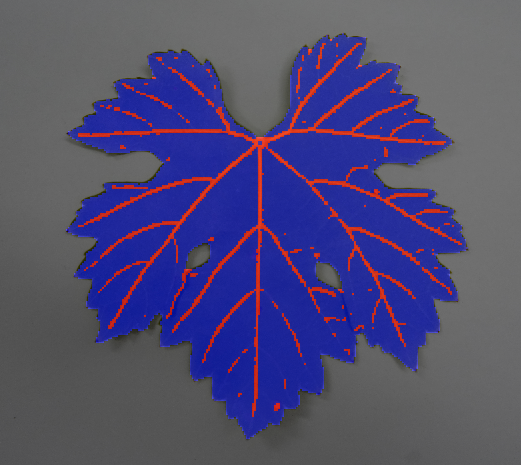}
                };
        \end{scope}
        
        \draw [flow] 
            (1.2, -0.5) 
            to[out=225,in=45] (0.7, -1)
            to[out=0,in=180] (3.1, -1)
            to[out=45,in=225] (add1);
        \draw [flow]    
            (3.9, -0.5)
            to[out=225,in=45] (3.3, -1.1)
            to[out=0,in=180] (5.9, -1.1)
            to[out=45,in=225] (add2);
        \draw [flow]    
            (6.75, -0.5)
            to[out=235,in=45] (6.3, -1.1)
            to[out=0,in=180] (8.7, -1.1)
            to[out=45,in=225] (add3);   
        \draw [flow]    
            (9.55, -0.5)
            to[out=225,in=45] (9.05, -1)
            to[out=0,in=180] (11.4, -1)
            to[out=45,in=225] (add4);     
        \draw [flow]    
            (12.2,-0.5)
            to[out=225,in=45] (11.7, -1)
            to[out=0,in=180] (13.8, -1)
            to[out=45,in=225] (add5);
        
        \draw [flow]
            (-1.25, -0.5) -- (-0.3, -0.5);
            
        \draw [flow]
            (0.6, -0.5) -- (1.35, -0.5);
            %\draw [flow]
            %    (2, -0.5) -- (2.1, -0.5);         
        \draw [flow]
            (3, -0.5) -- (add1.west);
        \draw [flow]
            (add1.east) -- (4, -0.5);
            %\draw [flow]
            %    (4.9, -0.5) -- (5, -0.5);       
        \draw [flow]
            (6, -0.5) -- (add2.west);
        \draw [flow]
            (add2.east) -- (6.75, -0.5);    
            %\draw [flow]
            %    (7.65, -0.5) -- (7.75, -0.5);          
        \draw [flow]
            (8.75, -0.5) -- (add3.west);
        \draw [flow]
            (add3.east) -- (9.6, -0.5);   
            %\draw [flow]
            %    (10.25, -0.5) -- (10.35, -0.5);            
        \draw [flow]
            (11.3, -0.5) -- (add4.west);
        \draw [flow]
            (add4.east) -- (12.2, -0.5);  
            %\draw [flow]
            %    (12.6, -0.5) -- (12.7, -0.5);            
        \draw [flow]
            (13.6, -0.5) -- (add5.west);
        
        \draw [flow]
            (add5.east) -- (14.7, -0.5);    
            
        \node[filters] at (0.75, 1.5) {\footnotesize{32}};
        \node[filters] at (2.75, 1.25) {\footnotesize{2x64}};
        \node[filters] at (5.25, 1) {\footnotesize{2x128}};
        \node[filters] at (8, 1) {\footnotesize{2x128}};
        \node[filters] at (10.95, 1.25) {\footnotesize{2x64}};
        \node[filters] at (13.5, 1.5) {\footnotesize{2x32}};

\end{tikzpicture}
    \caption{U-Net architecture (X-ception style)}
    \label{fig:xception_unet}
\end{figure}

The last approach (\(3^{rd}\)) includes, also, a U-Net like architecture with skipped connections inspired from the X-ception \citep{xception} architecture (Fig.\ref{fig:xception_unet}). This time the encoder/decoder consist of blocks of double Separable Convolutional layers of 32, 64 and 128 filters. In a typical U-Net model used for such tasks, an encoder's layer residual is added to the input of its symmetric decoder layer whereas in this case each layer's output is added to the next layer's output. 

\subsection{Unsupervised learning based semantic segmentation}

For Unsupervised Learning (UL) based image segmentation the model is not provided with the ground truth labels but instead learns by its self the distribution of pixel intensity levels in order to segment the image and group together pixels that share similar intensity values. This time the model is trained on a larger unlabeled dataset of 240 pictures without corresponding trimaps. The dataset is acquired from the original one (Section 3.1) by the use of image augmentation techniques (random rotation, random zoom etc.). The goal is to develop an UL based image segmentation architecture that will undertake segment labeling and relieve experts from the difficult and time-consuming process of annotation. The NN architecture is kept the same as before (UNet) but due to the lack of labeled targets, a different loss function is applied such as the one for the Fuzzy C-Means (FCM) \citep{besdek,besdek2} method:
\[
J_{FCM} = \sum_{j\in \Omega}\sum_{k=1}^{C}u_{jk}^q||y_j-\upsilon_k||^2
\]
where $u_{jk}$ is a membership function for the $j^{th}$ pixel in the $k^{th}$ cluster, $||y_j-\upsilon_k||$ is the distance of pixel value $y$ at location $j$ from the class centroid $\upsilon_k$, $C$ is the number of clusters and $\Omega$ is the spatial domain of the image. The parameter $q\in[1, \infty]$ is a weighting exponent on each fuzzy membership. Unlike K-Means clustering method where each pixel is assigned or not to a cluster, in FCM a pixel is assigned a membeship value for each cluster that can take any value in $[0, 1]$ as long as: $ \sum_{k=1}^Cu_{jk}=1, \forall j \in \Omega$. By assigning the softmax output of our model to the membership functions, as is done in \cite{rfcm}, we obtain a differentiable loss function that depends only on the statistics of pixel intensity values. It also has to be mentioned that the number of clusters $(C)$ is predefined and set before training.  

\section{Results}

To the best of our knowledge, up till now there exists no other study that addresses semantic segmentation of vine leaf phenotypic characteristics so there were no references available in the literature regarding optimum network architectures for the specific task or any results regarding accuracy and performance. 
%\subsection{Supervised segmentation}
\par
\begin{figure}[ht]
  \centering
  \subfigure{\includegraphics[width=0.32\columnwidth, height=50pt]{photos/result2.png}}
  \subfigure{\includegraphics[width=0.32\columnwidth, height=50pt]{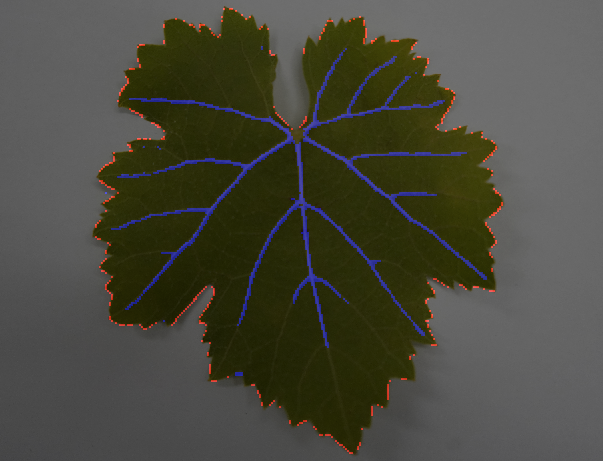}}
  \subfigure{\includegraphics[width=0.32\columnwidth, height=50pt]{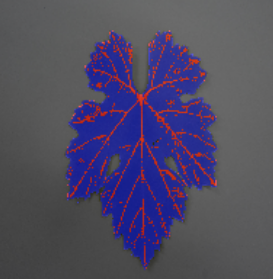}}
    \caption{SL based predictions for the 3 different approaches.}
  \label{fig:sl_prediction}
\end{figure}

Since this is a pixel-wise classification problem, class representation has to be taken into account. It is clear that the classes are imbalanced (indicatively: vein max width $\sim$7px for a 544x800 photo) which poses a challenge for developing unbiased, accurate predictive models. This was obvious in models' predictions: the models tend to predict well the blade leaf whereas fail to perform the same for leaf veins. In the majority of cases, veins were misclassified as part of the blade or as background (Fig. \ref{fig:sl_prediction}). For more downsized images (UNet-MobilNetV2) an opposite effect was observed were leaf blade pixels were misclassified as veins but in general image size had no major effect to results. To alleviate the problem, class weights inversely proportional to class size were introduced.
\par

\begin{comment}

\begin{figure}[t]
  \centering
  \includegraphics[width=0.32\columnwidth, height=60pt]{photos/result2.png}
  \hfill
  \includegraphics[width=0.32\columnwidth, height=60pt]{photos/predaugm.png}
  \hfill
  \includegraphics[width=0.32\columnwidth, height=60pt]{photos/pred6.png}
  \vfill
  \includegraphics[width=0.32\columnwidth, height=60pt]{photos/pred7.png}
  \hfill
  \includegraphics[width=0.32\columnwidth, height=60pt]{photos/pred8.png}
  \hfill
  \includegraphics[width=0.32\columnwidth, height=60pt]{photos/result3.png}
  %\includegraphics[width=0.24\columnwidth, height=42pt]{photos/trimap36.png}
  %\hfill
  %\includegraphics[width=0.24\columnwidth, height=42pt]{photos/trimap3850.png}
  %\hfill
  
  \caption{SL based predictions for the 3 different approaches.}
  \label{fig:sl_prediction}
\end{figure}

\end{comment}

\begin{table}[ht]
\caption{SL network performance results.}
 \tabcolsep=0.15cm
  \centering
    \begin{tabular}{|c|c||c|c|}
      \hline
    	\textit{Apr.} & \textbf{Model}  & \textbf{PA} & \textbf{MeanIoU} \\
      \hline
      \hline
      1 & \emph{UNet} & 0.94 & 0.74\\
      \hline
      2 & \emph{UNet(MobileNetV2-weights)} & 0.95 & 0.75\\
      \hline
      3 & \emph{UNet(Xception-like)} & 0.92 & 0.58\\
      \hline
    \end{tabular}
    \label{tab:SL-putcomes}
\end{table}

Our SL based models's performance reaches a plateau quite quickly showing satisfactory levels of Pixel Accuracy (PA) especially for the leaf veins, in contrast to other popular techniques like edge detection or thresholding which in addition require setting threshold values. For every model, we performed repeated runs experimenting on the hyperparameters and we concluded that validation set accuracy starts to rise around the $30^{th}$ epoch for training dataset sizes over 50 photos and 50 epochs were usually enough to yield satisfactory results. As an additional metric Intersection Over Union (IoU) was also used due to the fact that pixel accuracy can be misleading for imbalanced classification problems. Table \ref{tab:SL-putcomes} sum-ups the best training outcomes for all three approaches.

\begin{table}[ht]
\caption{UNet performance for increasing dataset sizes.}
 \tabcolsep=0.15cm
  \centering
    \begin{tabular}{|c|c||c|c|c|c|}
      \hline
    	\multirow{2}{*}{\emph{Apr.}} & \textbf{Dataset size}  & \multirow{2}{*}{\textbf{PA}} & \multicolumn{3}{c|}{\textbf{IoU}}\\
    	\cline{4-6}
    	 & \textbf{(train/valid/test)}  & & \textbf{Bkgr} & \textbf{Veins} & \textbf{Blade}\\    
      \hline
      \hline
      1 & \emph{12/4/4} & 0.73 & 0.62 & 0.012 & 0.0\\
      \hline
      2 & \emph{24/4/4} & 0.58 & 0.75 & 0.03 & 0.0\\
      \hline
      3 & \emph{36/5/4} & 0.72 & 0.62 & 0.0 & 0.0\\
      \hline
      4 & \emph{48/6/6} & 0.72 & 0.76 & 0.0 & 0.0\\
      \hline
      5 & \emph{60/8/8} & 0.96 & 0.98 & 0.28 & 0.88\\
      \hline
      6 & \emph{72/12/10} & 0.97 & 0.98 & 0.31 & 0.93\\
      \hline
    \end{tabular}
    \label{tab:dataset-results}
\end{table}

\begin{figure}[ht]
  \centering
  \subfigure[2 classes]{\label{fig:2-classes}\includegraphics[width=0.23\columnwidth, height=50pt]{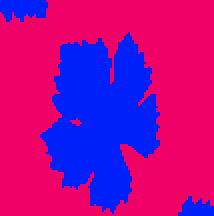}}
  \subfigure[3 classes]{\label{fig:3-classes}\includegraphics[width=0.23\columnwidth, height=50pt]{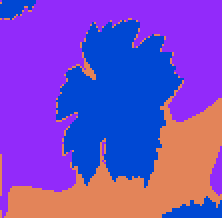}}
  %\vfill
  \subfigure[5 classes]{\label{fig:5-classes}\includegraphics[width=0.23\columnwidth, height=50pt]{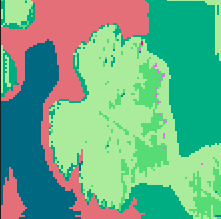}}
  \subfigure[10 classes]{\label{fig:10-classes}\includegraphics[width=0.23\columnwidth, height=50pt]{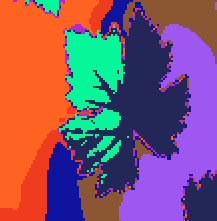}}
    \caption{UL based predictions.}
  \label{fig:UL-outcomes}
\end{figure}

%\subsection{Unsupervised segmentation}
Group accuracy of the UL based approach is low especially in cases where there is no intense color shade difference between objects. As can be seen in Fig.\ref{fig:UL-outcomes} the model manages to separate quite accurately the leaf blade from the background but fails to do the same for the leaf veins. This is clearly depicted in the metrics: experimenting for 3 classes (blade, leaf, background) the model achieves a very high IoU for leaf blade but underperforms for the veins (Table \ref{tab:UL-sumup}). Increasing the number of classes forces the model to segment the image into more groups and for classes $\geq$ 5 the veins start to stand out but are still inseparable from leaf blade regions (Fig.\ref{fig:UL-outcomes}).

\begin{figure}[ht]
  \centering
  \subfigure{\includegraphics[width=0.32\columnwidth, height=50pt]{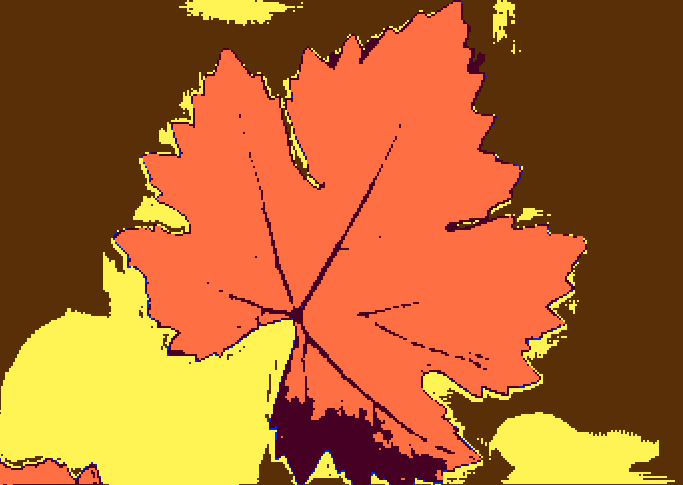}}
  \subfigure{\includegraphics[width=0.32\columnwidth, height=50pt]{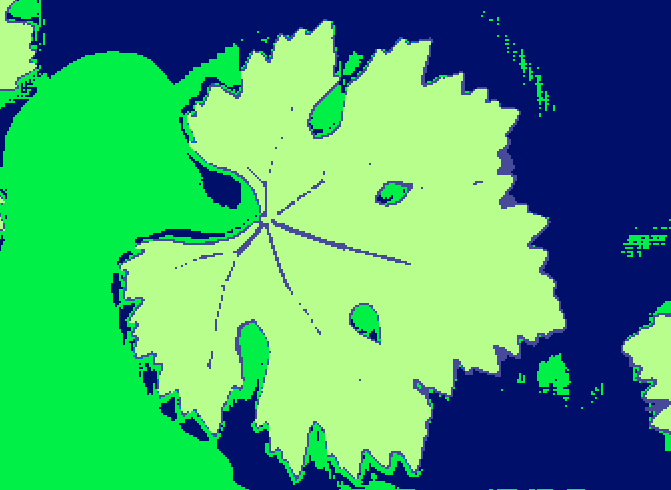}}
  \subfigure{\includegraphics[width=0.32\columnwidth, height=50pt]{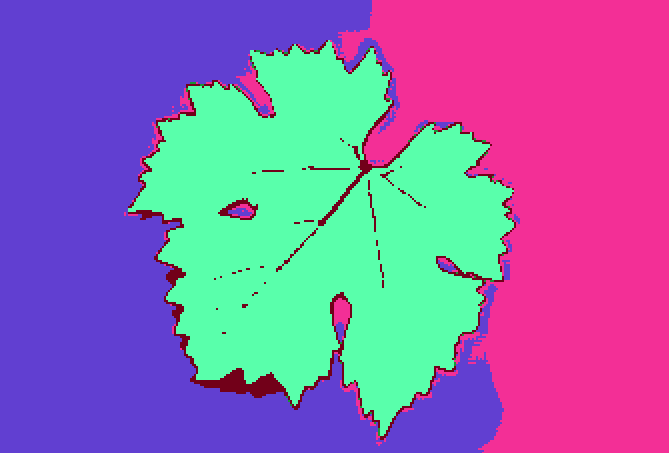}}
    \caption{UL results for contrast-enhanced images (classes=5).}
  \label{fig:sigmoid_correction}
\end{figure}

\begin{table}[ht]
\caption{Unsupervised network performance results (classes=3).}
 \tabcolsep=0.15cm
  \centering
    \begin{tabular}{|c|c||c|c|}
      \hline
    	\textit{} & \textbf{Region of interest}  & \textbf{IoU} & \textbf{MeanIoU} \\
      \hline
      \hline
      1 & \emph{Leaf blade} & 0.87 & \multirow{3}{*} {0.50}\\
      \cline{1-2} \cline{2-3}
      2 & \emph{Leaf veins} & 0.001 & \\
      \cline{1-2} \cline{2-3}
      3 & \emph{Background} & 0.57 & \\
      \hline
    \end{tabular}
    \label{tab:UL-sumup}
\end{table}
Low contrast areas e.g. leaf blade and vein endings pose a problem because they share similar pixel value intensities to the point that inferior lateral veins are indistinguishable from the blade.
The lack of feature richness is a common problem to image quality which can be mitigated by the application of image enhancement techniques \citep{hung}. In an effort to improve performance we applied sigmoid correction to the images in the preprocessing stage which seems to slightly improve overall performance (Fig.\ref{fig:sigmoid_correction}).

\section{Conclusion} 
In this study we introduced a new computer vision and ML challenging task, the vine leaf segmentation for precision agriculture purposes: insitu needs, laboratory analysis and robotics. We developed a vine leaf semantic segmentation framework that detects two basic phenotypic characteristics: leaf veins and leaf blade. In this direction we provided some indicative preliminary solutions. Our NN trained on images taken under controlled conditions, performed adequately and achieved promising levels of prediction accuracy even with a very small training dataset. It has to be stressed that this study is still ongoing and towards these directions we are developing a service that will automatically detect the leaf's characteristics and calculate quantities such as leaf width and length, number of superior lateral veins, blade area surface and vein to area ratio. The sparsity of well annotated data dictates the use of UL based semantic segmentation techniques and this is the direction our research interest is turning to. As a future work, apart from feature enhancement, we plan to tackle the low contrast regions problem by incorporating spatial and continuity constraints and asymmetric loss function variants e.g. focal loss. Furthermore, our interest focuses on developing a merging algorithm for over-segmented picture partitions that will improve segmentation accuracy.

\section*{Acknowledgements} \relax
We acknowledge support of this work by the project “AGRO4+" - Holistic approach to Agriculture 4.0 for new farmers” (MIS 5046239) which is implemented under the Action “Reinforcement of the Research and Innovation Infrastructure”, funded by the Operational Programme "Competitiveness, Entrepreneurship and Innovation" (NSRF 2014-2020) and co-financed by Greece and the European Union (European Regional Development Fund).

\bibliography{ifacconf.bib}

\end{document}